\documentclass{article}

\usepackage{orcidlink}     
\usepackage[affil-it]{authblk}

\usepackage{arxiv}
\usepackage{booktabs,siunitx,multirow,tabularx}
\usepackage{booktabs,siunitx,threeparttable}
\usepackage{subcaption}
\usepackage{tikz}
\usetikzlibrary{arrows.meta,positioning}
\usepackage{array}
\usepackage{xcolor}
\usepackage[utf8]{inputenc} 
\usepackage[T1]{fontenc}    
\usepackage{hyperref}       
\usepackage{url}            
\usepackage{booktabs}       
\usepackage{amsfonts}       
\usepackage{amsmath} 
\usepackage{nicefrac}       
\usepackage{microtype}      
\usepackage{lipsum}		
\usepackage{microtype} 
\usepackage[numbers]{natbib}
\usepackage{graphicx}
\usepackage{xcolor}
\usepackage{placeins} 
\usepackage{float} 
\usepackage{doi}
\usepackage{tabularx}

\definecolor{skyblue}{RGB}{0, 191, 245} 
\hypersetup{
    colorlinks=true,       
    citecolor=skyblue,        
    linkcolor=red,         
    urlcolor=magenta       
}

\title{Advancing Forest Fires Classification using Neurochaos Learning}

\author[1]{Kunal Kumar Pant\orcidlink{0009-0002-2748-8096}$^{\dagger}$}
\author[1]{Remya Ajai A S\orcidlink{0000-0002-7920-3374}$^{*,\dagger}$}
\author[2]{Nithin Nagaraj\orcidlink{0000-0003-0097-4131}}

\affil[1]{Department of Electronics and Communication Engineering, Amrita Vishwa Vidyapeetham, Kollam, Kerala, India} 
\affil[2]{Complex Systems Programme, National Institute of Advanced Studies (NIAS), IISc Campus, Bengaluru 560012, Karnataka, India} 
\date{} 

\renewcommand{\shorttitle}{\textit{Advancing Forest Fires Classification} }

\hypersetup{
pdftitle={Advancing Forest Fires Classification using Neurochaos Learning},
pdfsubject={q-bio.NC, q-bio.QM},
pdfauthor={Kunal K. Pant, Remya A S, Nithin Nagaraj},
pdfkeywords={Neurochaos Learning, Forest Fires, ChaosNet, Machine Learning, Deep Learning, Natural Disasters, Brain-Inspired Learning},
}

\makeatletter
\renewcommand\maketitle{%
  \begin{center}
    {\LARGE\bfseries \@title \par}%
    \vskip 1em
    {\AB@authlist \par}%
    \vskip .8em
    {\AB@affillist \par}%
  \end{center}%
}
\makeatother

\begin{document}
\maketitle

\begingroup
\renewcommand\thefootnote{}%
\footnotetext{*Corresponding author: \texttt{remya@am.amrita.edu}.}%
\addtocounter{footnote}{-1}%
\footnotetext{ $^\dagger$Equal contribution.}
\addtocounter{footnote}{-1}%

\endgroup

\begin{abstract}
    Forest fires are among the most dangerous and unpredictable natural disasters worldwide. Forest fire can be instigated by natural causes or by humans. They are devastating overall, and thus, many research efforts have been carried out to predict whether a fire can occur in an area given certain environmental variables. Many research works employ \textit{Machine Learning (ML) and Deep Learning (DL)} models for classification; however, their accuracy is merely adequate and falls short of expectations. This limit arises because these models are unable to depict the underlying nonlinearity in nature and extensively rely on substantial training data, which is hard to obtain. We propose using \textbf{Neurochaos Learning (NL)}, a chaos-based, brain-inspired learning algorithm for forest fire classification. Like our brains, NL needs less data to learn nonlinear patterns in the training data. It employs one-dimensional chaotic maps, namely the Generalized Lüroth Series (GLS), as neurons. NL yields comparable performance with ML and DL models, sometimes even surpassing them, particularly in  low-sample training regimes, and unlike deep neural networks, NL is interpretable as it preserves causal structures in the data. \textbf{Random Heterogenous Neurochaos Learning (RHNL)}, a type of NL where different chaotic neurons are randomnly located to mimic the randomness and heterogeneity of human brain gives the best F1 score of $1.0$ for the Algerian Forest Fires Dataset. Compared to other traditional ML classifiers considered,  RHNL also gives high precision score of $0.90$ for Canadian Forest Fires Dataset and $0.68$ for Portugal Forest Fires Dataset. The results obtained from this work indicate that Neurochaos Learning (NL)  architectures achieve better performance than conventional machine learning classifiers, highlighting their promise for developing more efficient and reliable forest fire detection systems.
\end{abstract}

\keywords{Neurochaos Learning \and Forest Fires \and ChaosNet \and Machine Learning \and Natural Disasters \and Brain-Inspired Learning}

\section{Introduction}
Forest fires are a major global concern. They have enormous environmental impacts, such as loss of vegetation, endangering wildlife, disrupting air quality, pollution of water resources, etc. The composition and structure of forests are also extensively influenced by the fire regime (Heinselman, $1973$; Wright and Bailey, $1982$, as cited in \citep{flannigan2000climate}). Frequency, seasonality, size,   type, intensity and severity are the six main components of forest fires. Study in \citep{rowell2000global} reports that there is a large amount of evidence that points to a trend of increase in forest fires (both in numbers and size). This results from the relationship between El Niño and climate change. Variations in winds and sea surface temperatures over the tropical Pacific Ocean give rise to a type of global climatic phenomenon known as El Niño or El Niño–Southern Oscillation (ENSO).  Evidence is mounting that the world is experiencing a positive feedback cycle whereby deforestation and forest fires made worse by climate change lead to a rise in the frequency of El Niño events, which in turn triggers more forest burning~\citep{cotter2009forest}. There may be an increase in El Niño's frequency and intensity, which means that the world will experience warmer and more severe weather, which may increase the number of forest fires. 

\begin{figure}[!ht]  
    \centering
    \includegraphics[width=0.8\textwidth]{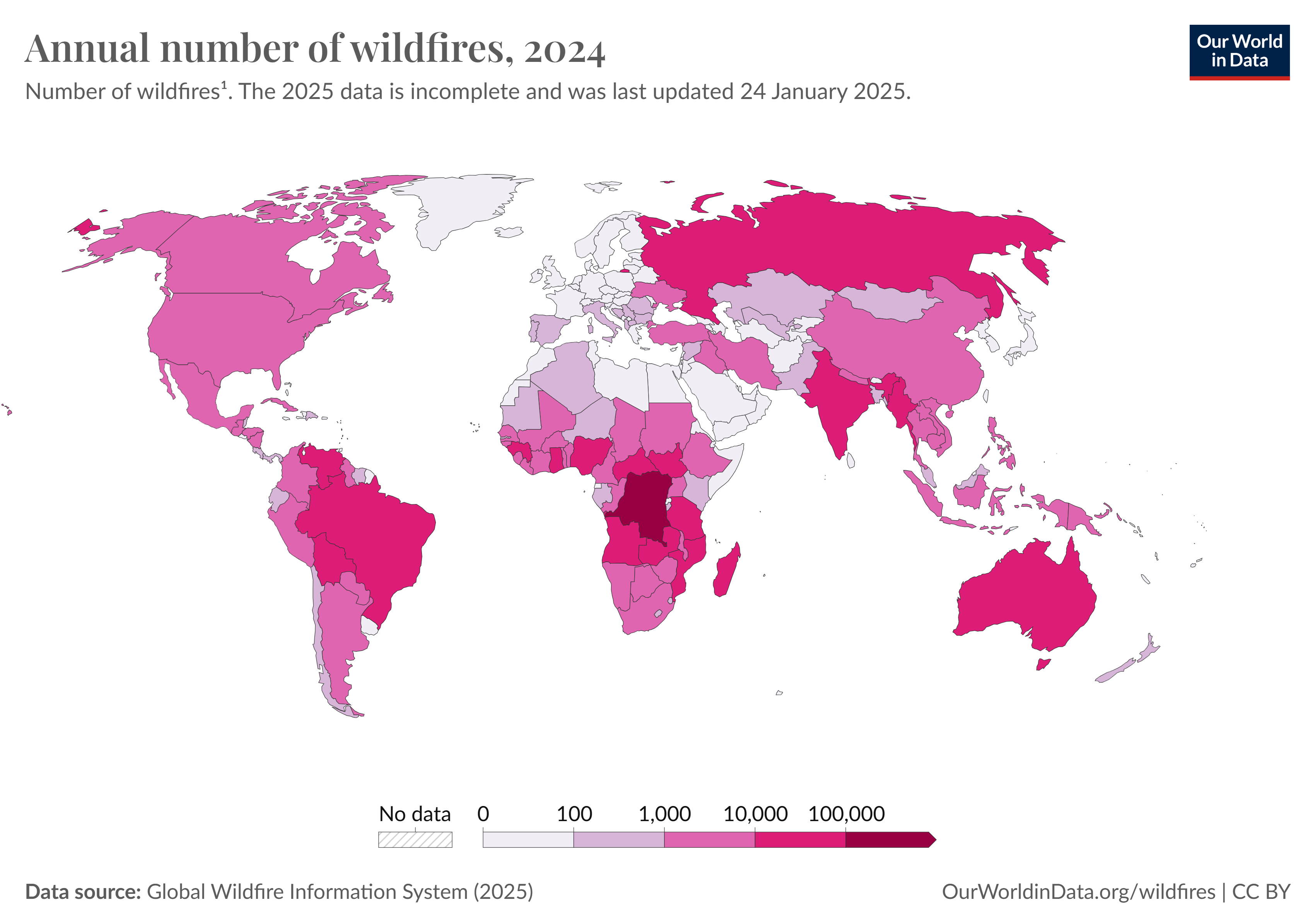}
    \caption{\rmfamily Annual number of wildfires ($2024$). Source: Our World in Data \citep{ourworldindata_wildfiresmap2024}, licensed under \href{https://creativecommons.org/licenses/by/4.0/}{CC BY $4.0$}.}
    \label{fig1:AnnualNumber of wildfires}
\end{figure}

\begin{figure}[!h] 
    \centering
    \includegraphics[width= 0.8\textwidth]{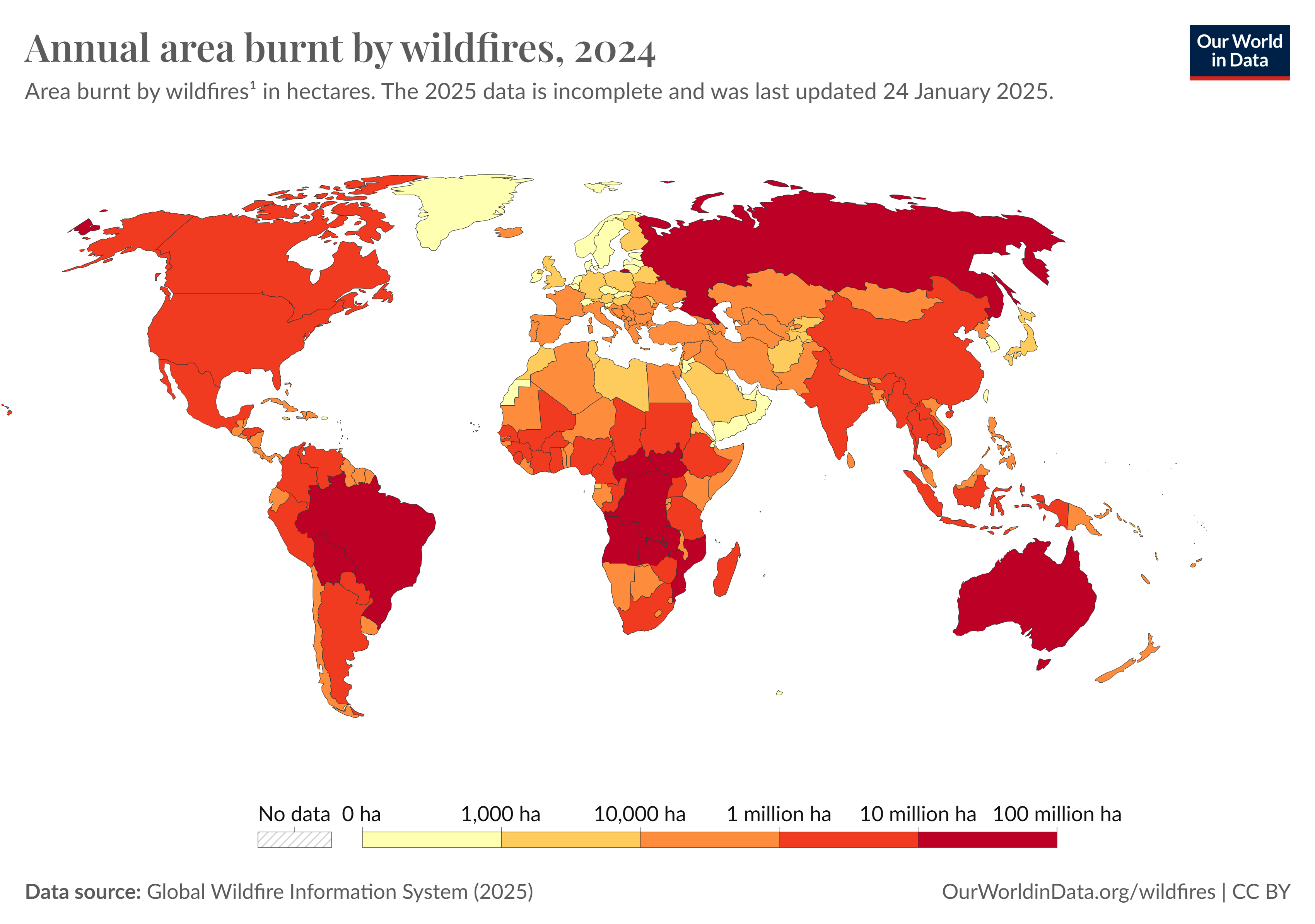
}
    \caption{\rmfamily Annual area burnt by wildfires ($2024$). Source: Our World in Data \citep{ourworldindata_wildfires2024}, licensed under \href{https://creativecommons.org/licenses/by/4.0/}{CC BY $4.0$}.}
    \label{fig2:Annual Area of wildfires}
\end{figure}

Each year, around 1.5 million square miles of land are impacted by fire, according to estimates from the European Space Agency~\citep{parsaforest}. To put this into perspective, this area is larger than India and nearly four times the size of Nigeria. Figure~\ref{fig1:AnnualNumber of wildfires} provides an overview of wildfire occurrences across the globe during 2024, and Figure~\ref{fig2:Annual Area of wildfires} depicts the total burnt area associated with these events. Collectively, these figures highlight that regions including Africa, South America, Russia, and Australia exhibited both elevated wildfire frequency and substantial landscape degradation. In India, the SNPP-VIIRS (Suomi National Polar-Orbiting Partnership - Visible Infrared Imaging Radiometer Suite) identified $3,45,989$ forest fires, whereas the MODIS (Moderate Resolution Imaging Spectroradiometer) sensor identified $52,785$ forest fires during the forest fire season between November $2020$ and June $2021$~\citep{fsi_ff_activities}. $35.71\%$ of India's forests have not yet experienced fires of any meaningful size, nevertheless $54.40\%$ of forests are subject to sporadic fires, $7.49\%$ to moderately regular fires, and $2.40\%$ to high incidence levels.

Numerous prior studies have shown that Machine Learning (ML) or even Deep Learning (DL) models demonstrate limited efficacy in dealing with natural adversities and anthropogenic issues. This is due to the inherent complexities and variability of natural systems. Even with stochastic and probabilistic elements incorporated into these existing models, they are not able to comprehend the intricacies of these natural systems completely. 
There is a need for models capable of integrating {\it Stochastic Resonance} into their architecture, which can process inputs that contain specific levels of noise, as opposed to relying solely on ideal noise-free inputs, which traditional Machine Learning (ML) and Deep Learning (DL) models do. Here, \textbf{ChaosNet} \citep{harikrishnan2021noise} can be leveraged. ChaosNet is a type of Neurochaos Learning (NL) ~\citep{harikrishnan2019novel} architecture -- a chaos based brain-inspired learning architecture. NL is inspired by the chaotic firing of biological neurons. NL has the flexibility of having a a classifier based on the cosine similarity measure or can be combined with classical ML classifiers~\citep{sethi2023neurochaos}. ChaosNet is a shallow neural network consisting of a single layer of chaotic 1D Generalised L\"{u}roth Series (GLS) maps as neurons~\citep{balakrishnan2019chaosnet}.

Random Heterogenous Neurochaos Learning (RHNL) architecture is an updated structure of NL where randomness and the  characteristics of neuronal heterogeneity in the human brain are taken into account together with chaotic behavior~\citep{as7random}. In RHNL, neurons based on GLS and the logistic map are positioned randomly within the input layer. There are three variations of RHNL structure namely $RHNL_{25L75G}$, $RHNL_{50L50G}$ and $RHNL_{75L25G}$. $RHNL_{25L75G}$ consists of $25\%$ of the input locations assigned to logistic map neurons, with the remaining occupied by GLS map neurons. In $RHNL_{50L50G}$, $50\%$ of the locations contain GLS map neurons, and the remaining positions are filled with logistic map neurons. $75\%$ locations in $RHNL_{75L25G}$ are randomly assigned with logistic map function and the remaining with GLS function. 

In this work, we focus on classification of forest fire occurrences as well as comparing the performance of traditional ML methods with NL. To carry out the same, three key regions were chosen: Algeria \citep{abid2020predicting}, Portugal \citep{cortez2007data}, and Canada \citep{sayad2019predictive}. These regions were selected due to the availability of well-documented forest fires datasets that had been thoroughly collected, efficiently pre-processed, and extensively tested across various models, enabling a comparative analysis of model performance. The acronyms used in this study are summarized in Table~\ref{tbl1} for ease of reference.

\begin{table}[ht]
\caption{\rmfamily Acronyms used throughout the paper.}\label{tbl1}
\centering
\renewcommand{\arraystretch}{1.2}
\setlength{\tabcolsep}{20pt} 
{\rmfamily
\begin{tabular}{|c|p{6cm}|c|}
\hline
\textbf{Item} & \textbf{Acronym} \\  
\hline
Neurochaos Learning & NL \\
\hline
Random Heterogeneous Neurochaos Learning & RHNL \\
\hline
$RHNL_{25L75G}$&{RHNL architecture comprises $25\%$of neurons based on the logistic map, with the remaining neurons employing the GLS map. Features are fed to cosine similarity classifier} \\
\hline
$RHNL_{25L75G}+SVM$ & $RHNL_{25L75G}$ with SVM classifer \\
\hline
$RHNL_{50L50G}$&{RHNL architecture comprises $50\%$of neurons based on the logistic map, with the remaining neurons employing the GLS map Features are fed to cosine similarity classifier} \\
\hline
$RHNL_{50L50G} +SVM$ & $RHNL_{50L50G}$ with SVM classifer \\
\hline
$RHNL_{75L25G}$&{RHNL architecture comprises $75\%$of neurons based on the logistic map, with the remaining neurons employing the GLS map. Features are fed to cosine similarity classifier} \\
\hline
$RHNL_{75L25G}+SVM$ & $RHNL_{75L25G}$ with SVM classifer \\
\hline
Deep Learning & DL \\
\hline
Decision Trees & DT \\
\hline
Random Forest & RF \\
\hline
European-Forest-Fire-Information System & EFFIS \\
\hline
Middle East and North Africa region & MENA \\
\hline
Geographic Information System & GIS \\
\hline
Artificial Neural Network & ANN \\
\hline
Fine Fuel Moisture Code & FFMC \\
\hline
Drought Code  & DC \\
\hline
Duff Moisture Code & DMC \\
\hline
Fire Weather Index & FWI \\
\hline
Initial Spread Index  & ISI \\
\hline
Buildup Index & BUI \\
\hline
Relative Humidity(\%) & Rh \\
\hline
Rain Fall(mm) & Rain  \\
\hline
Wind Speed(Km/Hr) & Ws \\
\hline
High Sample Training Regime & HSTR \\
\hline
Low Sample Training Regime & LSTR \\
\hline
Normalized Difference Vegetation Index & NDVI \\
\hline
Thermal Anomalies & TA  \\
\hline
Land Surface Temperature & LST \\
\hline

\end{tabular}
}
\end{table}

\section{Literature Review}
\subsection{Initial Developments: Fuel Models ($1960$ - $1980$s)}
Researchers have long sought to understand the characteristics of fire, leading to advancements in forest fire prediction. Early efforts, whether focusing on fire occurrence, spread, or effects of different types of fuels and environmental conditions, can be traced back to the $1940$s, as documented in \citep{fons1946analysis}. This study utilized controlled experimental environments to analyze the dynamics of fire spread, resulting in the development of mathematical models that provided theoretical frameworks for real-world applications. These models facilitated the prediction of key factors, such as the rate of spread, ignition time, and heat transfer processes. The study emphasized the growing necessity for empirical data to enhance the accuracy of forest fire predictions. The ideas and theories proposed in this were confirmed in~\citep{sanchez1967flame}.

The next most notable research was \citep{rothermel1972mathematical} in the $1970$s. It described the intricate relation of fire spread to variables like wind, fuel etc. and came up with equations and mathematical models for forecasting how fast and how intensely fire will spread through a continuous layer of fuel lying on the ground. It discussed about the conversion of a forest ground fire to a crown fire, given that sufficient heat is generated. After that, in the $1980$s, existing models were improved by developing tools to assist in selecting appropriate fuel models to predict fire behavior \citep{anderson1981aids}, improving the practical application of Rothermel's model.

\subsection{Integration of Remote Sensing and GIS ($1980$s - $2000$)}
While the initial use of remote sensing for forest fire mapping began in the $1960$s, the widespread adoption of remote sensing and GIS techniques emerged during the $1980$s. With the advancement in computer and data storage services, GIS came up as a new possibility. GIS makes it possible to manage loads of spatial information and derive different sorts of models like cartographic models. This is done by combining these layers of stored information in different ways. \citep{chuvieco1989application} takes on hazard mapping of forest fires. This study was centered on the Mediterranean coast of Spain.
 It implements this by using two methods: high-resolution imagery and test areas already affected by the forest fire. This makes it easier to test the efficacy of the hazard mapping. 

\subsection{Emergence of Machine Learning Techniques ($1990$s – Present, $2025$)}  
The rapid advancements in computational power and hardware during the $1990$s propelled the widespread adoption of machine learning techniques across various fields. These methods were soon applied to forest fire prediction, leveraging cartographic models generated from GIS data. Among the earliest studies to incorporate machine learning, \citep{blackard1999comparative} evaluated two approaches: Gaussian Discriminant Analysis and ANNs, a form of deep learning (with deep learning categorized under the broader domain of machine learning). ANNs were an early attempt by the scientific community to mimic the structural and functional principles of neurons in the human brain. However, it is important to acknowledge that current ANNs are far from accurately replicating the complex workings of the billions of neurons in the brain. As highlighted in \citep{baba2024neural}, ANNs do not precisely emulate biological neurons. Instead, artificial neurons are simplified representations that loosely approximate the behavior of their biological counterparts. Despite this, ANN models consistently outperformed Gaussian Discriminant Analysis in this  study~\citep{blackard1999comparative}. Other studies like \citep{stojanova2006learning} incorporated many other ML techniques like different types of decision trees, decision forests, regression techniques, etc.

However traditional ML algorithms lack in fully mimicking human brain in terms of important aspects such as chaotic behaviour, randomness and heterogeneity. For the first time in ~\citep{balakrishnan2019chaosnet}, authors proposed the novel Neurochaos Learning architecture, where the chaotic nature of brain is considered for developing the input layer of classification model. Later in ~\citep{as2023analysis}, the 1D GLS map used as neurons are replaced by 1D logistic map to analyse the classification performance. Two more properties of brain namely randomness and heterogeneity are considered and RHNL architecture is proposed in ~\citep{as7random}. RHNL gives better classification performance for various dataset considered and even outperforms DL in the low training sample regime.

\section{Dataset Description}
\subsection{Algerian Forest Fire Dataset (AFF)}
According to \citep{san2020advance}, Algeria is the most impacted nation among MENA countries and ranks fourth among all countries monitored by EFFIS.
The dataset considered in this study, sourced from \citep{abid2020predicting}, consists of observations gathered from two separate areas of Algeria: Béjaa and Sidi Bel-Abbès from the north-east and north-west regions, respectively. Data acquisition was carried out during the period from June to September $2012$, corresponding to the months identified as having the highest frequency of fire occurrences between $2007$ and $2018$. The dataset categorizes each observation into one of two classes — “fire” or “not fire” as it has been designed as a classification task. The total number of samples comes up to \textbf{$244$}. The exact class distribution is provided in Table \ref{AFF class distribution}.

\begin{table}[ht]
\centering
\renewcommand{\arraystretch}{1.2}
\caption{\rmfamily AFF Dataset Class Distribution.}
\label{AFF class distribution}
{\rmfamily
\begin{tabular}{|c|c|}
\hline
\textbf{Class} & \textbf{Occurences} \\
\hline
"not fire" $\longrightarrow$ $0$ & $106$ \\
\hline
"fire" $\longrightarrow$ $1$ & $138$ \\
\hline
\end{tabular}
}
\end{table}

The features utilized for classification include Temperature, Relative Humidity (RH), Rain, Wind Speed (Ws), and components of the Fire Weather Index (FWI) system. 
The FWI system also called the Canadian Forest Fires Weather Index system or the Forest Weather Index consists of $6$ components \citep{van1974structure}.  \textbf{Fine Fuel Moisture Code (FFMC)} is a numerical value that indicates the moisture content of cured fine fuels, such as litter, and weights approximately 0.05 pounds per square foot. \textbf{Duff Moisture Code (DMC)} is the amount of moisture in decomposed and loosely packed organic material which is specifically 2 to 4 inches deep and weighs around $1 lb/ft^{2}$ in a dry state. \textbf{Drought Code (DC)} accounts for the deep layer of compact and organic matter weighing around $10 lb/ft^{2}$ in a dry state. \textbf{Initial Spread Index (ISI)} is derived from a combination of the Fine Fuel Moisture Code (FFMC) and wind speed (Ws). It depicts the spread rate while not considering the impact of fuel quantities. \textbf{Buildup Index (BUI)} is the combination of DMC and DC. \textbf{Forest Weather Index (FWI)} is the combination of BUI and ISI.  A train-test split of 80-20 was used to partition the dataset for model training and performance evaluation.

\FloatBarrier

\subsection{Canadian Forest Fire Dataset (CFF)}
An experiment was developed to examine the constructed dataset with the aim of forecasting wildfire events in a designated forest region of Canada between 2013 and 2014. Fire zone information was sourced from the Canadian Wildland Fire Information System (CWFIS).
In this study \citep{sayad2019predictive}, data conversion was carried out using tools like GDAL (Geospatial Data Abstraction Library) and HEG (HDF-EOS to GeoTIFF Conversion Tool).
Since the data was extracted from satellite images, challenges such as geometric distortions and variations due to cloud cover and atmospheric conditions need to be considered. Atmospheric correction was applied, as the raw data had already undergone preprocessing for other aspects, making any other additional techniques like geo-referencing, ortho-rectification or radiometric correction unnecessary. The data then was clipped using spatiotemporal data interpolation and extrapolation techniques and tools. 

Three parameters as given in Table~\ref{tbl3}, were chosen based on three factors: the temperature of the soil, the health of the crop, and a fire indicator. The health of the crop and the soil temperature are important considerations if forecasting when heat or lightning will start wildfires. The third parameter Thermal Abnormalities (TA), gives  a detection confidence when a fire is sufficiently intense to be identified, and consequently, provides  direct information about the fire. An 80–20 train–test split was utilized to partition the dataset for training and validating the model’s performance.

\begin{table}[ht]
\centering
\caption{\rmfamily Parameters of the CFF dataset.}\label{tbl3}
\begin{tabular}{|l|l|l|}
\hline
\textbf{Parameter} & \textbf{Unabbreviated} & \textbf{Aspect} \\  
\hline
NDVI & Normalized Difference Vegetation Index & Crop's health \\
\hline
LST & Land Surface Temperature & Soil Temperature \\
\hline
TA & Thermal Anomalies & Fire Indicator \\
\hline
\end{tabular}
\end{table}

\subsection{Portugal Forest Fire Dataset(PFF)}
The Montesinho Natural Park in Portugal's Tras-os-Montes Northeast region is where the Portugal Forest Fire data is sourced from January $2000$ to December $2003$. 
An inspector collected part of the  data by registering several features such as time, date, and spatial location every time a forest fire occurred within a $9$x$9$ grid. Along with the different elements of the FWI system, the type of vegetation involved and the total burned area were also recorded.
 A polytechnic institute in Braganc gathered the remaining portion of the database, which included many weather observations. A meteorological station in the heart of Moesinho Park captured these throughout a $30$ minute period. Inserted within a supra-Mediterranean climate, the average annual temperature range was from $8^\circ\text{C} -12^\circ\text{C}$~\citep{cortez2007data}. The attributes of PFF  dataset considered are FFMC, DMC, DC, ISI, temperature, relative humidity, wind, rain and area.

\FloatBarrier
\section{Classifiers}  
The three Forest Fire datasets namely AFF, CFF and PFF are classified using traditional ML models such as Support Vector Machine (SVM), Random Forest (RF), Decision Tree (DT), XGBoost and latest proposed models such as NL and RHNL. We tune all traditional ML models via grid search with scoring (macro F1-score) and 5-fold cross-validation, selecting the hyperparameters that maximize the macro-F1 score across folds. For NL models~\citep{harikrishnan2019novel,sethi2023neurochaos,as7random}, the three hyperparameters we need to tune are discrimination threshold ($b$), initial neural activity ($q$) and noise intensity ($\epsilon$). Hyperparameters were optimized using five-fold cross-validation to achieve the best model performance. Table~\ref{Reference:MLHyper} provides an overview of the set of hyperparameters tuned for all the ML and NL algorithms.

\begin{table}[htb]
\caption{\rmfamily References corresponding to the tuned hyperparameters for each algorithm.}\label{Reference:MLHyper}
\centering
\renewcommand{\arraystretch}{1.2}
{\rmfamily
\begin{tabularx}{\linewidth}{|l|X|}
\hline
\textbf{Algorithm} & \textbf{Hyperparameters tuned} \\  
\hline
SVM & $C, kernel, gamma, degree$ \\
\hline
RF & $min\_samples\_leaf$, $max\_depth$, $n\_estimators$, $min\_samples\_split$ \\
\hline
DT & $max\_depth$, $min\_samples\_split$ $min\_samples\_leaf$ \\
\hline
DT with AdaboostM1 & $learning\_rate$, $n\_estimators$ \\
\hline
DT with Bagging & $max\_samples$, $n\_estimators$, $max\_features$\\
\hline
XGBoost & $learning\_rate$, $colsample\_bytree$,  $n\_estimators$, $max\_depth$, $reg\_alpha$, $subsample$,  $reg\_lambda$ \\
\hline
NL&$initial$  $neural$ $activity(q)$, $discrimination$ $threshold(q)$, $noise$ $intensity(\epsilon)$\\
\hline
\end{tabularx}
}
\end{table}

\subsection{Support Vector Machine (SVM)}
The hyperparameters for SVM~\citep{boser1992training}, a support vector machine for classification based tasks are \textit{kernel}, \textit{C}, \textit{degree}, and \textit{gamma}. \textit{C}, also known as the regularization parameter, controls the bias-variance trade-off of the algorithm. \textit{kernel} is the feature mapping choice which defines the decision boundary shape. \textit{Gamma} influences the radius of a single training point and \textit{degree} is the polynomial degree for the poly kernel. Table~\ref{SVM_Hyper} gives the values of various parameters considered for SVM.

\begin{table}[htb]
\caption{\rmfamily Parameter grid for SVM.}\label{SVM_Hyper}
\centering
\renewcommand{\arraystretch}{1.2}
{\rmfamily
\begin{tabularx}{\linewidth}{|l|X|}
\hline
\textbf{Hyperparameter} & \textbf{Grid} \\  
\hline
$C$ &  $[0.1,1,10,100]$ \\
\hline
$kernel$ & $['rbf','sigmoid', 'linear','poly']$ \\
\hline
$gamma$ &  $[1,0.1,0.01,0.001]$ if kernel is $['rbf',sigmoid', 'poly']$
                            else $['scale']$\\
\hline
$degree$ & $[2,3,4]$ if kernel ==$'poly'$ else $[3]$ \\
\hline
\end{tabularx}
}
\end{table}
\FloatBarrier

\subsection{Random Forest (RF)}
The hyperparameters for the RF model \citep{breiman2001random},  include \textit{max\_depth}, \textit{min\_samples\_leaf}, \textit{min\_samples\_split}, \textit{n\_estimators}. \textit{n\_estimators} defines the number of trees which are being grown in a forest. \textit{max\_depth} is the maximum allowed depth each tree in the forest is allowed. \textit{min\_samples\_leaf} declares the minimum samples required in a leaf and \textit{min\_samples\_split} defines the minimum samples to split an internal node. The various parameters used for configuring the Random Forest classifier are summarized in Table~\ref{RF_Hyper}.

\begin{table}[ht]
\caption{\rmfamily Parameter grid for RF.}
\label{RF_Hyper}
\centering
\renewcommand{\arraystretch}{1.2}
{\rmfamily
\begin{tabular}{|c|c|}
\hline
\textbf{Hyperparameter} & \textbf{Grid} \\  
\hline
$max\_depth$ & $[50,40,30,20,10]$ \\
\hline
$min\_samples\_split$ & $[10,5,2]$ \\
\hline
$min\_samples\_leaf$ & $[4, 2, 1]$ \\
\hline
$n\_estimators$ & $[250, 200,150,100,50]$ \\
\hline
\end{tabular}
}
\end{table}

\subsection{Decision Tree (DT)}
The hyperparameters for the DT model \citep{quinlan1986induction}, include \textit{max\_depth}, which is used to define the maximum depth to which a tree can grow, \textit{min\_samples\_leaf}, which declares the minimum number of samples required for a leaf (node), and \textit{min\_samples\_split}, which defines the minimum samples to split in an internal node. Table~\ref{DT_Hyper} summarizes the parameters employed in configuring the DT classifier.
 
\begin{table}[ht]
\caption{\rmfamily Parameter grid for DT.}\label{DT_Hyper}
\centering
\renewcommand{\arraystretch}{1.2}
{\rmfamily
\begin{tabular}{|c|c|}
\hline
\textbf{Hyperparameter} & \textbf{Grid} \\  
\hline
$max\_depth$ & $[10, 20, 30, 40, 50]$ \\
\hline
$min\_samples\_split$ & $[2,5,10]$ \\
\hline
$min\_samples\_leaf$ & $[1,2,4]$ \\
\hline
\end{tabular}
}
\end{table}

\subsection{DT with Adaboost M1}
The hyperparameters for the boosted model \citep{schapire2013explaining}, include \textit{learning\_rate}, which helps shrink eaech weak learner's contribution and \textit{n\_estimators}, which accounts for the number of boosting rounds. The parameters utilized for configuring the DT with Adaboost M1 classifier are summarized in Table~\ref{Adaboost_Hyper}.

\begin{table}[htbp]
\caption{\rmfamily Parameter grid for DT with Adaboost M1.}\label{Adaboost_Hyper}
\centering
\renewcommand{\arraystretch}{1.2}
{\rmfamily
\begin{tabular}{|c|c|}
\hline
\textbf{Hyperparameter} & \textbf{Grid} \\  
\hline
$learning\_rate$ &  $[.01, .1, .5, 1.0]$ \\
\hline
$n\_estimators$ &  $[50, 100, 150, 200]$ \\
\hline
\end{tabular}
}
\end{table}
\FloatBarrier
\subsection{DT with Bagging}
The hyperparameters for the bagging model \citep{breiman1996bagging} include \textit{max\_features} which accounts for the number of features sampled per base estimator, \textit{max\_samples} which is the fraction of training samples drawn per base estimator, and \textit{n\_estimators}, which is the number of bootstrapped models to aggregate. The configuration parameters for the Decision Tree (DT) with the Bagging M1 classifier are detailed in Table~\ref{Bagging_Hyper}.

\begin{table}[htpb]
\caption{\rmfamily Parameter grid for bagging model.}\label{Bagging_Hyper}
\centering
\renewcommand{\arraystretch}{1.2}
{\rmfamily
\begin{tabular}{|c|c|}
\hline
\textbf{Hyperparameter} & \textbf{Grid} \\  
\hline
$max\_samples$ &  $[.6,.8,1.0]$ \\
\hline
$max\_features$ & $[.6,.8,1.0]$\\
\hline
$n\_estimators$ &  $[50, 100, 150, 200]$ \\
\hline
\end{tabular}
}
\end{table}

\FloatBarrier

\subsection{XGBoost}
 The XGBoost classifier\citep{chen2016xgboost} was utilized for binary classification, with parameter tuning focused on mitigating class imbalance and enhancing model performance. The hyperparameters for XGBoost in classification tasks include \textit{n\_estimators} which is the number of boosting rounds (trees), \textit{max\_depth} or the maximum tree depth, \textit{learning\_rate} which is the step size shrinkage per boosting step, \textit{colsample\_bytree} which defines the fraction of features sampled for each tree, \textit{subsample} which is the fraction of rows sampled per tree, \textit{reg\_alpha} for L1 regularization, and \textit{reg\_lambda} for L2 regularization. The configuration parameters for the XGBoost classifier are detailed in Table~\ref{XGBOOST_Hyper}.

\begin{table}[ht]
\caption{\rmfamily Parameter grid for XGBoost.}\label{XGBOOST_Hyper}
\centering
\renewcommand{\arraystretch}{1.2}
{\rmfamily
\begin{tabular}{|c|c|}
\hline
\textbf{Hyperparameter} & \textbf{Grid} \\  
\hline
$learning\_rate$ &  $[.01, .1, .2]$ \\
\hline
$n\_estimators$ &  $[50, 100, 150, 200] $\\
\hline
$max\_depth$ & $[3, 5, 7, 10]$ \\
\hline
$colsample\_bytree$ & $[.6, .8, 1.0]$ \\
\hline
$subsample$ & $[.6, .8, 1.0]$ \\
\hline
$reg\_alpha$ & $[0, .1, 1]$ \\
\hline
$reg\_lambda$ & $[1, 1.5, 2]$ \\
\hline
\end{tabular}
}
\end{table}

\subsection{ChaosNet}
Neurochaos Learning (NL) architecture~\citep{harikrishnan2019novel} represents a chaos-driven neuronal framework that emulates the intrinsic chaotic behavior of the human brain within its neural structure, a characteristic typically absent in traditional machine learning models. ChaosNet is the basic Neurochaos Learning (NL) architecture that consists of a layer of chaotic neurons, modeled as 1D Generalized Lüroth Series (GLS) maps. The hyperparameters defined for ChaosNet are initial neural activity ($q$), discrimination threshold ($b$) and $\epsilon$, the noise intensity. Initial neural activity is considered as the point from which the neuron starts firing when an input stimulus triggers it. The chaotic firing of each GLS neuron ceases once its activity value evolving from the initial neural state $q$ enters the $\epsilon$ neighborhood of the specific input stimulus. Hence, different neurons may stop firing at different times. The range of ChaosNet hyperparameters considered  are given in Table~\ref{NL parameter grid}.

\begin{table}[htbp]
\caption{\rmfamily Parameter grid for ChaosNet.}\label{NL parameter grid}
\centering
\renewcommand{\arraystretch}{1.2}
{\rmfamily
\begin{tabular}{|c|c|}
\hline
\textbf{Hyperparameter} & \textbf{Grid Range} \\  
\hline
$q$ &  $[.001 - 1.0]$\\
\hline
$b$ &   $[.01 -  0.5]$\\
\hline
$\epsilon$ & $[.001 - .5]$  \\
\hline
\end{tabular}
}
\end{table}
\FloatBarrier

\textbf{Random Heterogenous Neurochaos Learning (RHNL)} architecture is an updated structure of NL where randomness and heterogeneity properties of the human brain are considered along with chaotic behavior~\citep{as7random}. In RHNL, GLS and logistic map neurons are placed in the input layer at random locations.

There are three variations of RHNL structure namely $RHNL_{25L75G}$, $RHNL_{50L50G}$ and $RHNL_{75L25G}$. $RHNL_{25L75G}$ consist of $25\%$ locations with  logistic map neurons and remaining with GLS map neurons. In $RHNL_{50L50G}$, $50\%$ of locations are placed with logistic map neurons and remaining with GLS map neurons. $75\%$ locations in $RHNL_{75L25G}$ are randomnly assigned with logistic map function and the remaining with GLS function. 

In RHNL~\citep{as7random}, the features generated from input samples namely Firing Rate, Firing Time, Energy and Entropy are fed to cosine similarity classifier. The structure of RHNL where cosine similarity classifier in the output layer is replaced by SVM are termed as  $RHNL_{RH25L75G}$+SVM, $RHNL_{RH50L50G}$+SVM and $RHNL_{RH75L25G}$+SVM.
The range of values used for RHNL hyperparameters tuning  are given in Table~\ref{RHNL parameter grid}.

\begin{table}[htbp]
\caption{\rmfamily Parameter grid for RHNL.}\label{RHNL parameter grid}
\centering
\renewcommand{\arraystretch}{1.2}
{\rmfamily
\begin{tabular}{|c|c|}
\hline
\textbf{Hyperparameter} & \textbf{Grid} \\  
\hline
$q$ &  $[.001-.5]$\\
\hline
$b$ &   $[.01-.5]$\\
\hline
$\epsilon$ & $[.001-.3]$  \\
\hline
\end{tabular}
}
\end{table}
\FloatBarrier


\section{Results}
The performance of various classifiers namely SVM, RF, DT, DT with Adaboost M1,DT with Bagging, XGBoost, ChaosNet and RHNL are analysed for three different forest fire datasets namely AFF, CFF and PFF. Analysis were performed for both the High Sample Training Regime (\textbf{HSTR}) and Low Sample Training Regime (\textbf{LSTR}). In the low sample training regime, $100$ random and independent trials have been considered for training from $1,2,...,10$ data instances in each class.  

\subsection{Algerian Forest Fire}

\FloatBarrier

\begin{figure}[htpb]
    \centering
    \includegraphics[width=0.7\linewidth]
    {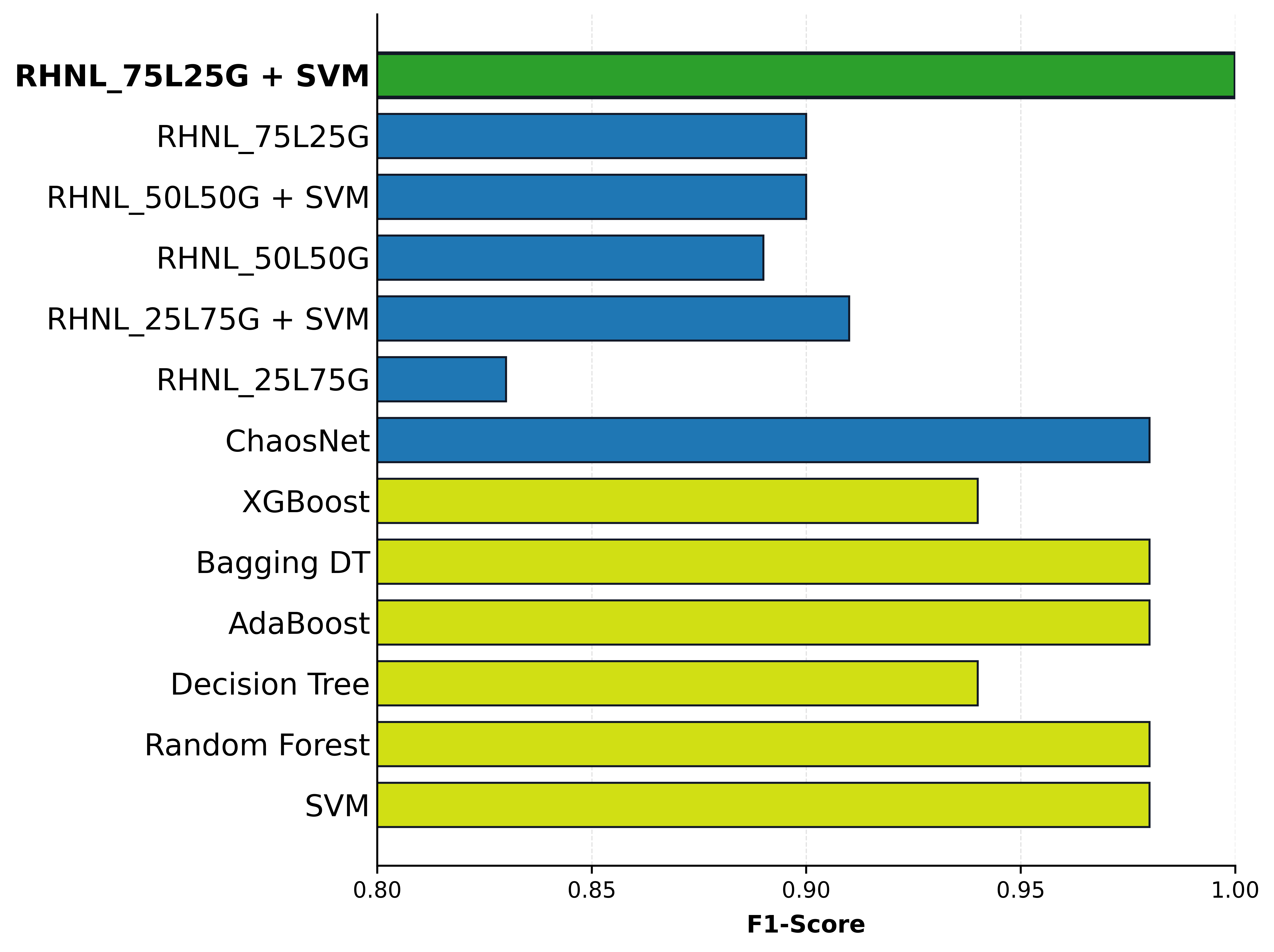}
    \caption{\rmfamily F1 Score obtained for AFF Dataset.}
    \label{AFF:fig:HSTR:f1}
\end{figure}

\begin{figure}[htpb]
    \centering
    \includegraphics[width=0.6\linewidth]
    {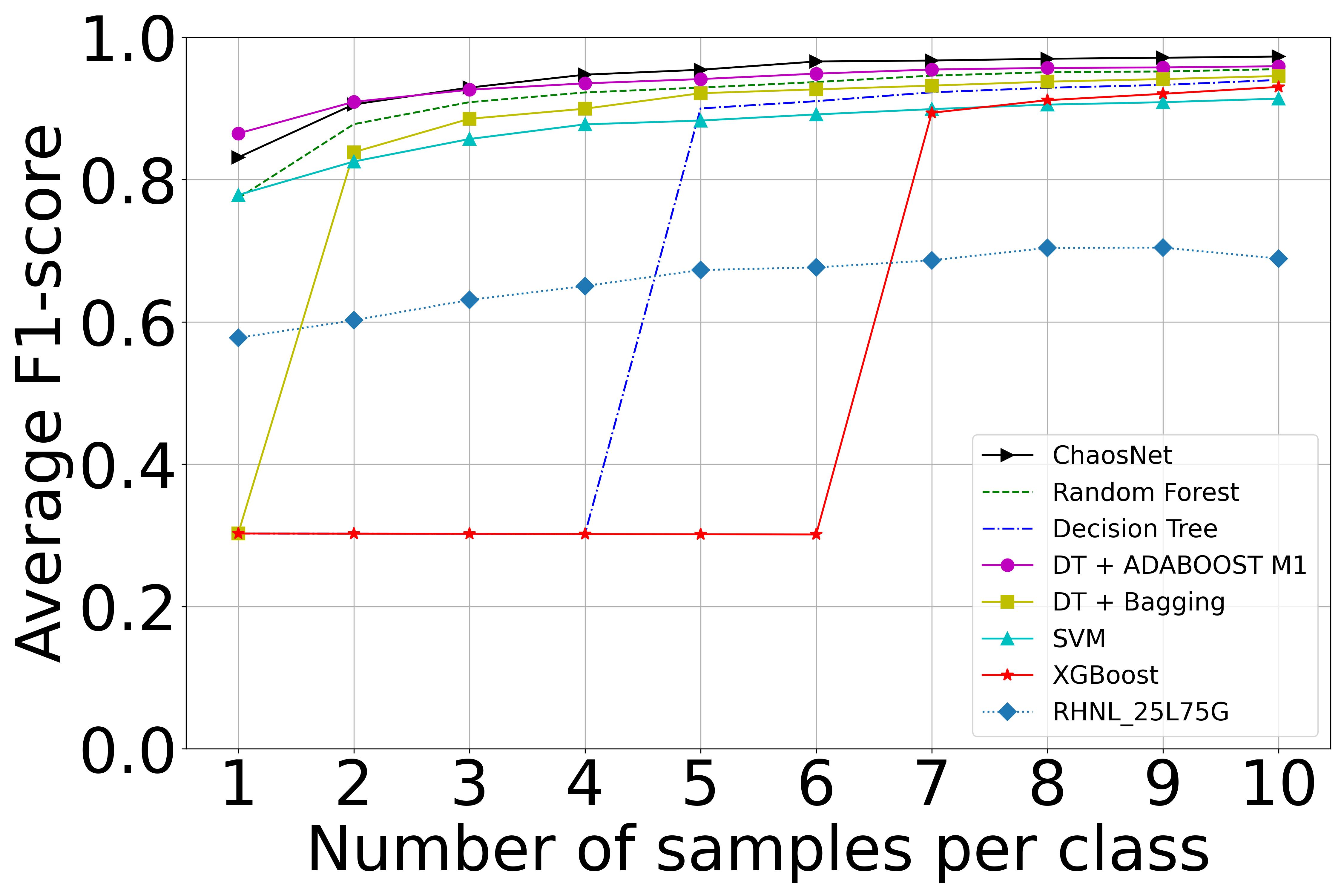}
    \caption{ \rmfamily \textbf{LSTR performance obtained for AFF Dataset}: Macro-F1 vs. samples.}
    \label{AFF:fig:LSTR}
\end{figure}

Classification of AFF dataset is done with all the considered classifiers. The hyperparameter tuned for AFF is given in Table~\ref{AFF:tbl:Best_Hyper}. The high sample training regime performance comparisons for various ML and NL models applied on the AFF dataset are provided in Table~\ref{AFF:tbl:Performance_Metrics}. As shown in Figure~\ref{AFF:fig:HSTR:f1}, the $RHNL_{75L25G}+SVM$ model achieved the highest F1-score of 1.0. Figure~\ref{AFF:fig:LSTR} depicts the comparative performance of all models in LSTR, clearly indicating the superior performance of ChaosNet over the others. From both the results, it can be noted that NL outperforms all of the baseline ML models for AFF dataset considered.

\begin{table*}[htbp]
\caption{\rmfamily Best Hyperparameter Values tuned for AFF Dataset.}
\label{AFF:tbl:Best_Hyper}
\centering
\renewcommand{\arraystretch}{1.3} 
\setlength{\tabcolsep}{34pt} 
\rmfamily{
\begin{tabularx}{\textwidth}{|>{\centering\arraybackslash}p{3.5cm}|>{\centering\arraybackslash}X|}
\hline
\textbf{Classifier} & \textbf{Best Hyperparameters} \\
\hline
SVM & $Kernel=Linear$, $C=10$, $Degree=3$, $Gamma=scale$ \\ 
\hline
RF & 
$max\_depth=10$, $min\_samples\_leaf= 1$,$min\_samples\_split=2$, $n\_estimators=50$ \\
\hline
DT& 
$max\_depth=10$, $min\_samples\_split= 10$, $min\_samples\_leaf= 1$ \\
\hline
\mbox{DT with Adaboost M1} & $learning\_rate=0.01$, $n\_estimators= 50$, $algorithm=$ $'SAMME'$ \\
\hline
DT with Bagging & $max\_features= 0.6$, $max\_samples=0.6$, $n\_estimators=50$ \\
\hline
XGBoost &
$scale\_pos\_weight=106/138$, $colsample\_bytree=0.6$, $learning\_rate=0.01$, $max\_depth= 3$, $n\_estimators=50$, $reg\_alpha= 0$, $reg\_lambda=1$, $subsample=0.8$ \\
\hline
ChaosNet & 
$q=.93$, $b=.49$, $\epsilon$=$.165$–$.167$  \\
\hline
$RHNL_{25L75G}$ & 
$q=.040$, $b=.199$, $\epsilon$= $.054$ 
\\
\hline
$RHNL_{50L50G}$ & 
$q=.067$, $b=.120$, $\epsilon$= $.105$ 
\\
\hline
$RHNL_{75L25G}$ & 
$q=.01$, $b=.21$, $\epsilon$= $.161$ 
\\
\hline
\end{tabularx}
}
\end{table*}

\begin{table*}[htbp]
\caption{\rmfamily Performance Metrics of Different Classification Models on AFF Dataset.}
\label{AFF:tbl:Performance_Metrics}
\centering
\renewcommand{\arraystretch}{1.3} 
\setlength{\tabcolsep}{12pt} 
\rmfamily{
\begin{tabularx}{\textwidth}{|>{\centering\arraybackslash}p{3.5cm}|>{\centering\arraybackslash}X|>{\centering\arraybackslash}X|>{\centering\arraybackslash}X|>{\centering\arraybackslash}X|}
\hline
\textbf{Classifier} & \textbf{F1 Score} & \textbf{Accuracy} & \textbf{Precision} & \textbf{Recall} \\
\hline
SVM &  0.98 &  0.98 & 0.98 &  0.98 \\
\hline
RF &  0.98 &  0.98 &  0.98 &  0.98 \\
\hline
DT & 0.94 & 0.94 & 0.95 & 0.93\\
\hline
DT with Adaboost M1 & 0.98 &  0.98 &  0.98 &  0.98 \\
\hline
DT with Bagging &  0.98 &  0.98 &  0.98 &  0.98 \\
\hline
XGBoost & 0.94 & 0.94 & 0.95 & 0.93 \\
\hline
ChaosNet &  0.98 & 0.98 & 0.98 &  0.98 \\
\hline
$RHNL_{25L75G}$ & 0.83 &  0.83 & 0.83 &  0.84 \\
\hline

$RHNL_{25L75G}+SVM$ & 0.91 &  0.92 & 0.92 &  0.91 \\
\hline
$RHNL_{50L50G}$ & 0.89 &  0.90 & 0.90 &  0.89 \\
\hline

$RHNL_{50L50G}+SVM$ & 0.98 &   0.98 &  0.98 &  0.98 \\
\hline
$RHNL_{75L25G}$ & 0.90 &  0.90 & 0.90 &  0.91 \\
\hline

$RHNL_{75L25G}+SVM$ & \bfseries 1.0 &  \bfseries 1.0 & \bfseries 1.0 &  \bfseries 1.0 \\
\hline
\end{tabularx}
}
\end{table*}

\FloatBarrier
\subsection{Canadian Forest Fires}
\begin{figure}[htpb]
    \centering
    \includegraphics[width=0.7\linewidth]
    {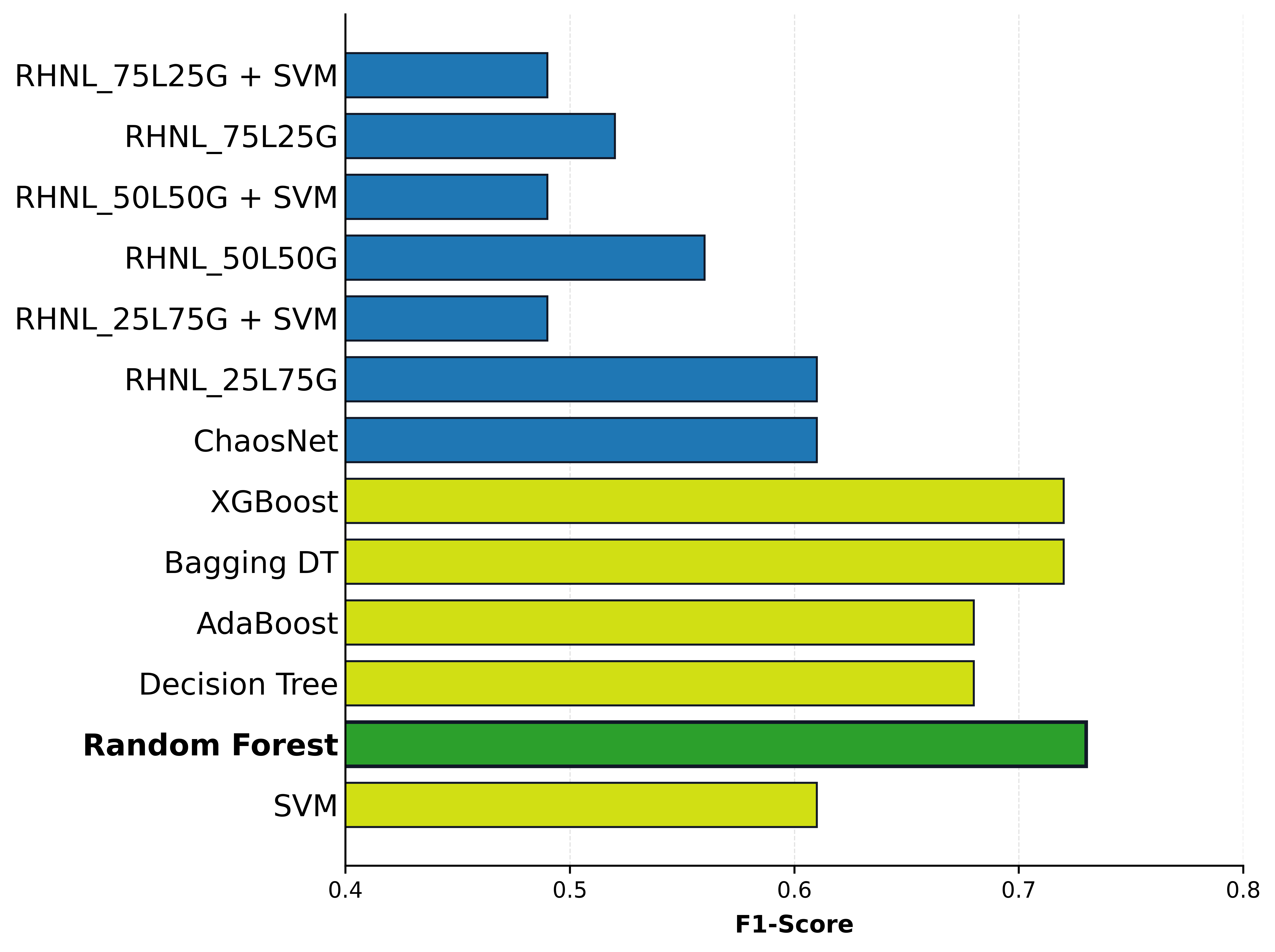}
    \caption{\rmfamily F1 Score obtained for CFF Dataset.}
    \label{CFF:fig:HSTR:f1}
\end{figure}

The hyperparameter adjustment for all respective models is represented in Table~\ref{CFF:tbl:Best_Hyper}. High sample training regime performance comparisons for applied ML and NL models are provided in Table~\ref{CFF:tbl:Performance_Metrics} and Figure~\ref{CFF:fig:HSTR:f1}. RF Classifier is giving high performance in terms of F1 Score, Accuracy and Recall.  But $RHNL_{50L50G}+SVM$ gives a high value of 0.90 for precision. This indicates that majority of the positive samples of forest fire are correctly classified with RHNL compared to other classifiers.

\begin{figure}[htpb]
    \centering
    \includegraphics[width=0.6\linewidth]
    {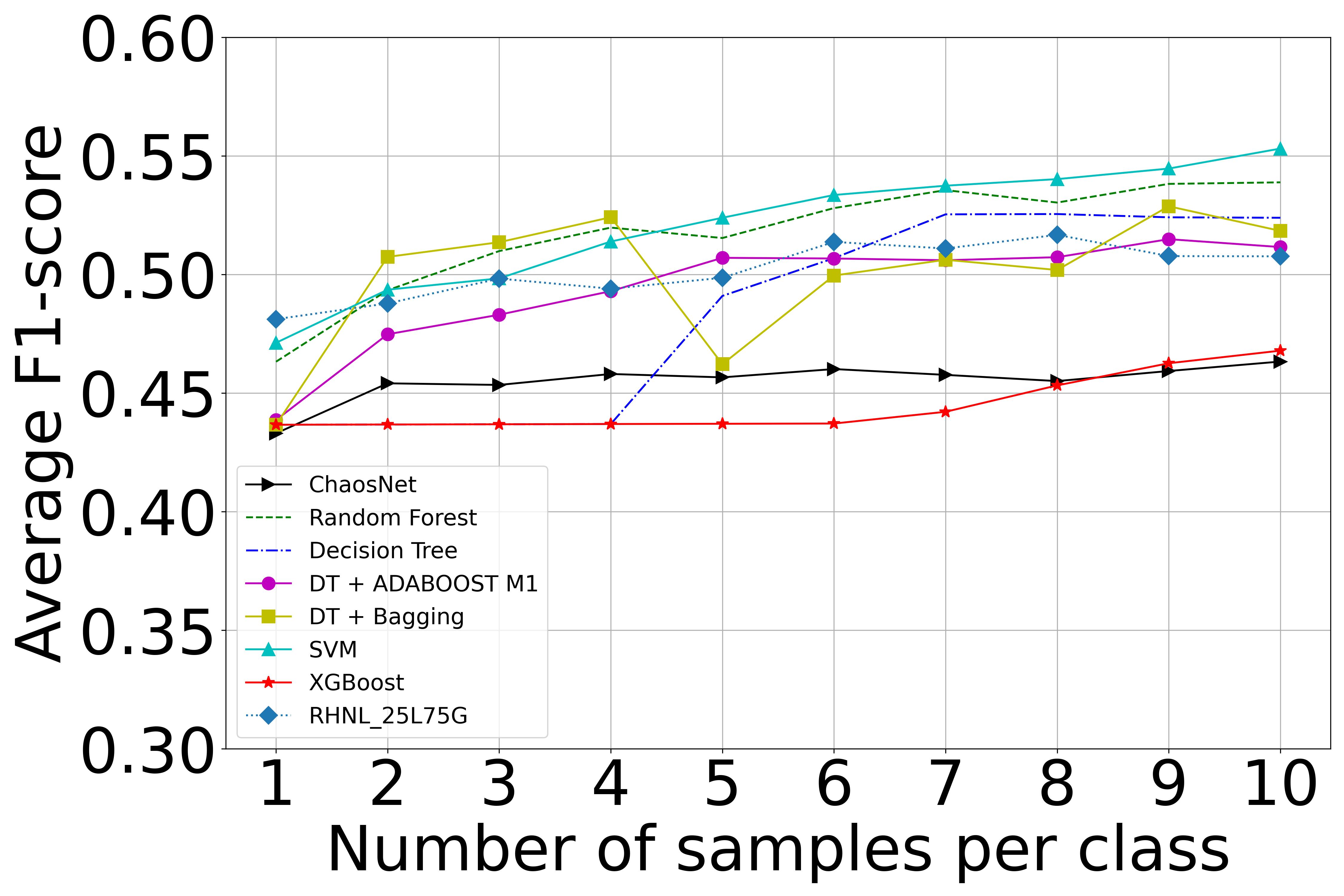}
    \caption{ \rmfamily \textbf{LSTR performance obtained for CFF Dataset}: Macro-F1 vs. samples.}
    \label{CFF:fig:LSTR}
\end{figure}

Fig~\ref{CFF:fig:LSTR} shows the performance of all 
models in a low sample training regime. SVM performs better with training samples from 4 to 10. However, RHNL outperforms all other classifiers with 1 training sample.

\begin{table*}[htb]
\caption{\rmfamily Best Hyperparameter Values for CFF Dataset.}
\label{CFF:tbl:Best_Hyper}
\centering
\renewcommand{\arraystretch}{1.3} 
\setlength{\tabcolsep}{12pt} 
\rmfamily{
\begin{tabularx}{\textwidth}{|>{\centering\arraybackslash}p{3.5cm}|>{\centering\arraybackslash}X|}
\hline
\textbf{Classifier} & \textbf{Best Hyperparameters} \\
\hline
SVM & 
$Kernel=RBF$, $C=100$, $Degree=3$, $Gamma=1$ \\
\hline
RF & 
$max\_depth=30$, $min\_samples\_split=2$, $min\_samples\_leaf= 2$,  $n\_estimators=250$  \\
\hline
Decision Tree(DT) & 
$max\_depth=20$, $min\_samples\_leaf= 1$, $min\_samples\_split=2$  \\
\hline
DT with Adaboost M1 & 
$learning\_rate=0.01$, $n\_estimators= 50$, $algorithm$= $'SAMME'$   \\
\hline
DT with Bagging & $max\_features= 1.0$, $max\_samples=0.8$, $n\_estimators=200$ \\
\hline
XGBoost &
$scale\_pos\_weight= 1$, $colsample\_bytree=0.8$, $learning\_rate=0.2$, $max\_depth= 10$, $n\_estimators=200$, $reg\_alpha= 0$, $reg\_lambda= 2$, $subsample= 0.8$ \\
\hline
ChaosNet & 
$q=.141$, $b=.499$, $\epsilon=.496$  \\

\hline
$RHNL_{25L75G}$ & 
$q=.344$, $b=.303$, $\epsilon$= $.261$ 
\\
\hline
$RHNL_{50L50G}$ & 
$q=.0344$, $b=.230$, $\epsilon$= $.261$ 
\\
\hline
$RHNL_{75L25G}$ & 
$q=.140$, $b=.489$, $\epsilon$=$.021$ \\
\hline
\end{tabularx}
}
\end{table*}

\vspace*{-2mm}

\begin{table*}[htb]
\caption{\rmfamily Performance Metrics of Different Models on CFF Dataset. Best values are highlighed in Bold font.}
\label{CFF:tbl:Performance_Metrics}
\centering
\renewcommand{\arraystretch}{1.3} 
\setlength{\tabcolsep}{12pt} 
\rmfamily{
\begin{tabularx}{\textwidth}{|>{\centering\arraybackslash}p{3.5cm}|>{\centering\arraybackslash}X|>{\centering\arraybackslash}X|>{\centering\arraybackslash}X|>{\centering\arraybackslash}X|}
\hline
\textbf{Classifier} & \textbf{F1 Score} & \textbf{Accuracy} & \textbf{Precision} & \textbf{Recall} \\
\hline
SVM & 0.61 & 0.67 & 0.62 & 0.67 \\
\hline
RF &  \bfseries 0.73 &  \bfseries 0.83 & 0.75 &  \bfseries 0.72\\
\hline
DT & 0.68 & 0.78 & 0.68 & 0.68  \\
\hline
DT with Adaboost M1 & 0.68 & 0.78 & 0.68 & 0.69  \\
\hline
DT with Bagging & 0.72 & \bfseries 0.83 & 0.76 & 0.69  \\
\hline
XGBoost & 0.72 & 0.81 & 0.73 & 0.71 \\
\hline
ChaosNet & 0.61 & 0.68 & 0.61 & 0.64 \\
\hline
$RHNL_{25L75G}$ & 0.61 &  0.74 & 0.60 &  0.62 \\
\hline

$RHNL_{25L75G}+SVM$ & 0.49 &  0.78 & 0.79 &  0.52 \\
\hline
$RHNL_{50L50G}$ & 0.56 &  0.59 & 0.60 &  0.65 \\
\hline

$RHNL_{50L50G}+SVM$ &  0.49 &   0.79 & \bfseries 0.90 &  0.53 \\
\hline
$RHNL_{75L25G}$ & 0.52 &  0.75 & 0.56 &  0.56 \\
\hline

$RHNL_{75L25G}+SVM$ &  0.49 &   0.79 &  0.89 &   0.53 \\
\hline
\end{tabularx}
}
\end{table*}

\FloatBarrier

\subsection{Portugal Forest Fires}

\begin{figure}[htpb]
    \centering
    \includegraphics[width=0.7\linewidth]
    {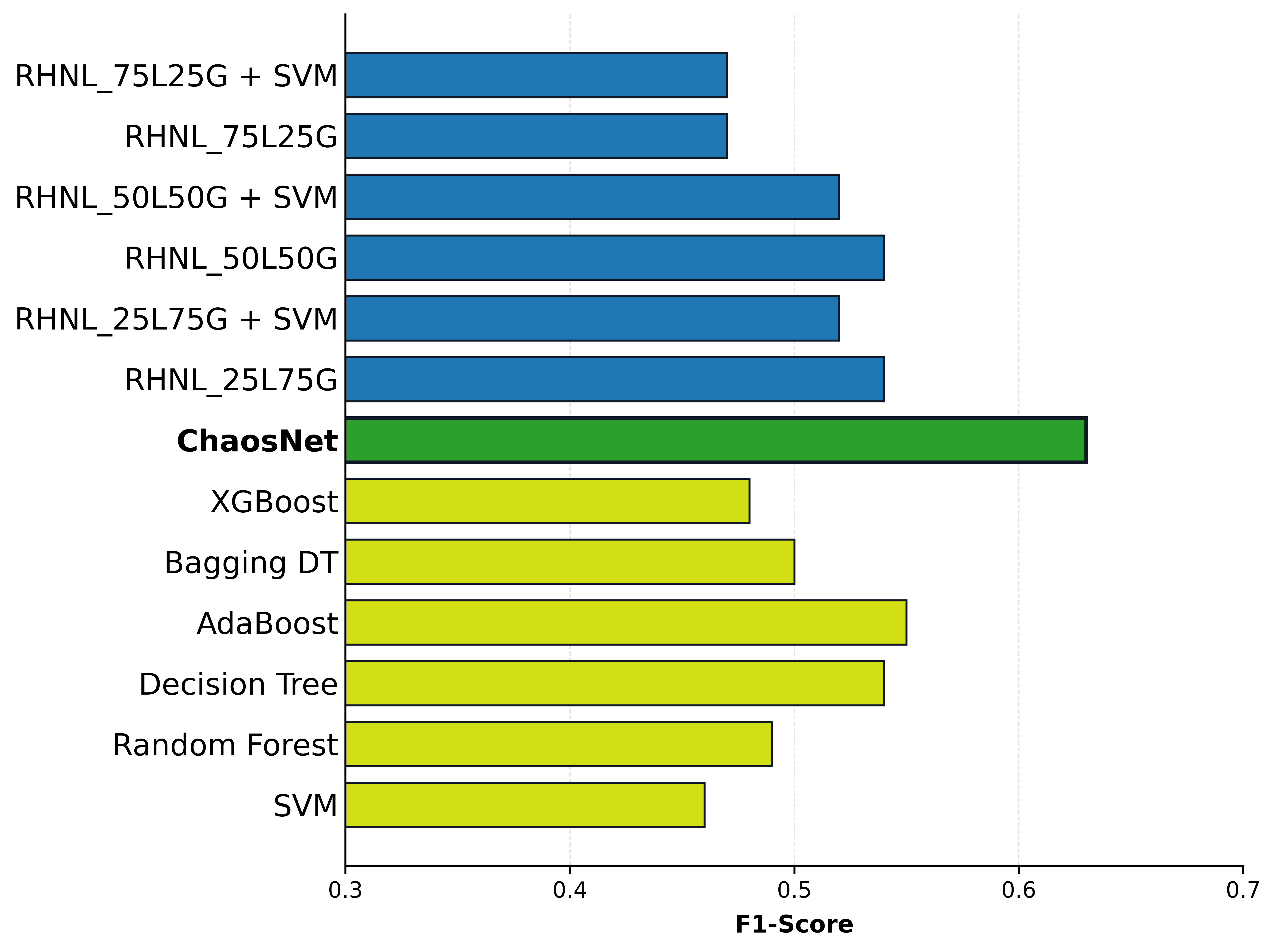}
    \caption{\rmfamily F1 Score obtained for PFF Dataset.}
    \label{PFF:fig:HSTR:f1}
\end{figure}

\begin{figure}[htpb]
    \centering
    \includegraphics[width=0.6\linewidth]
    {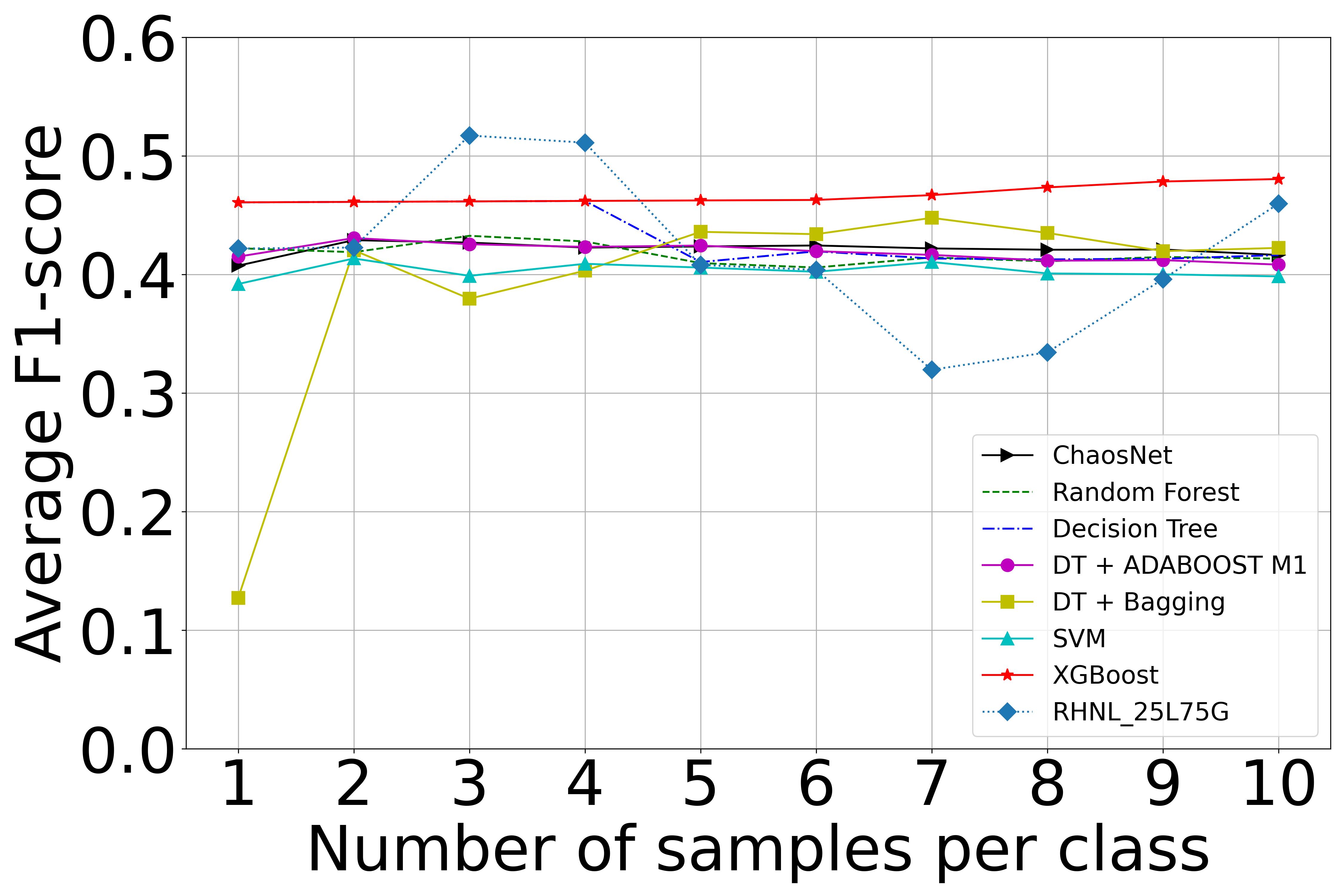}
    \caption{ \rmfamily \textbf{LSTR Performance obtained for PFF Dataset}: Macro-F1 vs. samples.}
    \label{PFF:fig:LSTR}
\end{figure}

The hyperparameter tuning Table~\ref{PFF:tbl:Best_Hyper} and the high sample training regime performance comparisons for various ML and NL models are provided in Table~\ref{PFF:tbl:HSTR}. Analysis shows that NL outperforms other traditional ML classifiers in terms of F1 Score, Precision and Recall. As depicted in Figure~
\ref{PFF:fig:HSTR:f1}, ChaosNet gives the highest F1 score of $0.63$ compared to all other classification models considered in HSTR. Figure~\ref{PFF:fig:LSTR} presents the comparative results in the low-sample training regime, where RHNL achieves the highest accuracy among all classifiers for training samples of three and four.

\begin{table}[ht]
\caption{\rmfamily Best Hyperparameter values for PFF dataset.}
\label{PFF:tbl:Best_Hyper}
\centering
\small
\renewcommand{\arraystretch}{1.35}
\setlength{\tabcolsep}{12pt} 
{\rmfamily

\begin{tabularx}{\linewidth}{|l|X|}
\hline
\textbf{Classifier} & \textbf{Best Hyperparameter Configuration} \\
\hline
SVM & $Kernel=sigmoid$, $C=1$, $Degree=3$, $\gamma=0.001$ \\
\hline
RF&
$n\_estimators=100$, $max\_depth=10$, $min\_samples\_split=5$,

$min\_samples\_leaf=1$ \\
\hline
DT& $max\_depth = 20$, $min\_samples\_split = 2$, $min\_samples\_leaf = 1$\\
\hline
DT with Adaboost M1 & $learning\_rate = 0.01$, $n\_estimators = 50$ \\
\hline
DT with Bagging& $max\_features = 1.0$, $max\_samples = 1.0$, $n_estimators = 200$\\
\hline
XGBoost & $Scale\_pos\_weight = 1$, ~$Colsample\_bytree = 0.6$, ~$Learning\_rate = 0.01$,
~$Max\_depth = 3$,~$N\_estimators = 50$,~$Reg\_alpha = 0$,~$Reg\_lambda = 1$,
~$Subsample = 0.8$ \\
\hline
ChaosNet &
$q =  .93$, $b = .49$, $\epsilon = .013$ \\
\hline
$RHNL_{25L75G}$ & 
$q = .123$, $b =  .028$, $\epsilon$ = $.031$ 
\\
\hline
$RHNL_{50L50G}$ & 
$q = .140$, $b =  .489$, $\epsilon$ =  $.021$ 
\\
\hline
$RHNL_{75L25G}$ & 
$q =  .020$, $b =  .219$, $\epsilon$ =  $.081$ 
\\
\hline
\end{tabularx}
}
\end{table}

\begin{table*}[ht]
\caption{\rmfamily Performance Comparison of different models on PFF Dataset. Best values are highlighed in Bold font.}\label{PFF:tbl:HSTR}
\centering
\renewcommand{\arraystretch}{1.35}
\setlength{\tabcolsep}{12pt} %
{\rmfamily
\begin{tabularx}{\textwidth}{|>{\centering\arraybackslash}p{3.5cm}|>{\centering\arraybackslash}X|>{\centering\arraybackslash}X|>{\centering\arraybackslash}X|>{\centering\arraybackslash}X|}
\hline
\textbf{Algorithm} & \textbf{F1 Score} & \textbf{Accuracy} & \textbf{Precision} & \textbf{Recall} \\
\hline
SVM & 0.46 & \bfseries 0.86 & 0.43 & 0.50 \\
\hline
RF & 0.49 & 0.81 & 0.50 & 0.50 \\
\hline
DT & 0.54 & 0.78 & 0.54 & 0.54 \\
\hline
DT with Adaboost M1 & 0.55 & 0.80 & 0.56 & 0.55  \\
\hline
DT with Bagging & 0.50 & 0.82 & 0.51 & 0.50  \\
\hline
XGBoost & 0.48 & 0.79 & 0.48 & 0.49  \\
\hline
ChaosNet & \bfseries 0.63 & 0.75 & 0.62 & \bfseries 0.69  \\
\hline
$RHNL_{25L75G}$ & 0.54 &  0.83 & 0.55 &  0.58 \\
\hline

$RHNL_{25L75G}+SVM$ & 0.52 &  0.85 & \bfseries 0.68 &  0.53 \\
\hline
$RHNL_{50L50G}$ & 0.54 &  0.83 & 0.55 &  0.58 \\
\hline

$RHNL_{50L50G}+SVM$ & 0.52 &  0.85 & \bfseries 0.68 &  0.53 \\
\hline
$RHNL_{75L25G}$ & 0.47 &  0.85 & 0.47 &  0.50 \\
\hline

$RHNL_{75L25G}+SVM$ &  0.47 &  0.85 & 0.43 &  0.50 \\ 
\hline
\end{tabularx}
}
\end{table*}

\FloatBarrier

\section{Conclusion and Future Work}
Forest fires classification continues to be a complex and challenging task in environmental data analysis due to the highly dynamic, nonlinear, and unpredictable behavior of fire spread. Accurate and timely classification of forest fire events is crucial for minimizing environmental damage, supporting proactive disaster management strategies, preserving biodiversity, and enabling efficient allocation of emergency resources.
In this study, we evaluated the effectiveness of various ML models alongside brain-inspired interpretable machine learning algorithms (ChaosNet, RHNL and hybrid NL$+$ML models) on multiple forest fire datasets. These datasets were either inherently classification-based (Algerian and Canadian datasets) or transformed into a classification format (Portugal dataset). For the High Sample Training Regime (HSTR), the Random Heterogeneous Neurochaos Learning (RHNL) model achieved 100\% classification accuracy, surpassing all traditional ML classifiers. Moreover, Neurochaos Learning (NL) demonstrated strong performance even under low training sample conditions, which is particularly significant in forest fire classification where data scarcity and irregular event occurrences often limit the availability of extensive training datasets.

Due to the inherent uniqueness and flexibility of NL, it can be incorporated into various chaos-based hybrid ML models or even chaos-based hybrid DL models and that approach can be leveraged further. The performance of other RHNL+ML classifiers can be further  analysed to enhance Forest Fire classification, particularly under limited training data conditions. In future work, this study will be extended to include forest fire datasets from other geographical regions to evaluate the robustness and adaptability of the developed classifiers under diverse environmental conditions. Incorporating region-specific climatic, vegetation, and topographical features will help in assessing the generalization capability of the proposed approach. Further investigation will focus on optimizing model parameters and exploring combining NL and/or chaos-based features with ensemble and deep learning–based techniques to improve prediction accuracy. Integrating satellite-based real-time information and temporal patterns can further enhance the model’s early detection efficiency and its practical applicability in forest fire monitoring systems.

\section{Acknowledgments}  
The authors gratefully acknowledge Amrita Vishwa Vidyapeetham, Amritapuri Campus, for providing the computational infrastructure and support for this work.


\begin{thebibliography}{10}
\expandafter\ifx\csname url\endcsname\relax
  \def\url#1{\texttt{#1}}\fi
\expandafter\ifx\csname urlprefix\endcsname\relax\def\urlprefix{URL }\fi
\expandafter\ifx\csname href\endcsname\relax
  \def\href#1#2{#2} \def\path#1{#1}\fi

\bibitem{flannigan2000climate} Flannigan, Michael D and Stocks, Brian J and Wotton, B Mike, Climate change and forest fires, Science of the total environment,volume262,
  number3, pages 221--229,2000,Elsevier

\bibitem{rowell2000global} Rowell, Andrew and Moore, Peter F and others, Global review of forest fires, 2000, Citeseer.

\bibitem{cotter2009forest} Cotter, Janet, Forest Fires: Influences of climate change and human activity, 2009, Greenpeace Research Laboratories Technical Note 05/2009.

\bibitem{parsaforest} Parsa, Peerzada Shuaib Amin and Zehra, Kaneez, Forest Fires and Climate Change: Causes, Effects and Management, Disaster Development, 107.

\bibitem{fsi_ff_activities} Forest Survey of India, Forest Fire Activities, Annual Report, 2025, 45--50.

\bibitem{ourworldindata_wildfiresmap2024} Hannah Ritchie and Max Roser, Annual Number of Wildfires Map, 2024, https://ourworldindata.org/grapher/annual-number-of-fires, Licensed under CC BY 4.0, Accessed: [24/01/25].

\bibitem{ourworldindata_wildfires2024} Hannah Ritchie and Max Roser, The annual area burnt by wildfires, 2024, https://ourworldindata.org/wildfires, Licensed under CC BY 4.0, Accessed: [24/01/25].


\bibitem{harikrishnan2021noise} Harikrishnan, Nellippallil Balakrishnan and Nagaraj, Nithin, When noise meets chaos: Stochastic resonance in neurochaos learning, Neural Networks, 143, 425--435, 2021, Elsevier.

\bibitem{harikrishnan2019novel} Harikrishnan, Nellippallil Balakrishnan and Nagaraj, Nithin, A novel chaos theory inspired neuronal architecture, 2019 Global Conference for Advancement in Technology (GCAT), 1--6, 2019, IEEE.

\bibitem{sethi2023neurochaos} Sethi, Deeksha and Nagaraj, Nithin and Harikrishnan, Nellippallil Balakrishnan, Neurochaos feature transformation for Machine Learning, Integration, 90, 157--162, 2023, Elsevier.

\bibitem{balakrishnan2019chaosnet} Balakrishnan, Harikrishnan Nellippallil and Kathpalia, Aditi and Saha, Snehanshu and Nagaraj, Nithin, ChaosNet: A chaos based artificial neural network architecture for classification, Chaos: An Interdisciplinary Journal of Nonlinear Science, 29(11), 2019, AIP Publishing.

\bibitem{as7random} AS, Remya Ajai and Nagaraj, Nithin, Random Heterogeneous Neurochaos Learning Architecture for Data Classification, Chaos Theory and Applications, 7(1), 10--30, Akif Akgül.

\bibitem{abid2020predicting} Abid, Faroudja and Izeboudjen, Nouma, Predicting forest fire in Algeria using data mining techniques: Case study of the decision tree algorithm, Advanced Intelligent Systems for Sustainable Development (AI2SD’2019) Volume 4-Advanced Intelligent Systems for Applied Computing Sciences, 363--370, 2020, Springer.

\bibitem{cortez2007data} Cortez, Paulo and Morais, Aníbal de Jesus Raimundo, A data mining approach to predict forest fires using meteorological data, 2007, Associação Portuguesa para a Inteligência Artificial (APPIA).

\bibitem{sayad2019predictive} Sayad, Younes Oulad and Mousannif, Hajar and Al Moatassime, Hassan, Predictive modeling of wildfires: A new dataset and machine learning approach, Fire Safety Journal, 104, 130--146, 2019, Elsevier.

\bibitem{fons1946analysis} Fons, Wallace L, Analysis of fire spread in light forest fuels, Journal of Agricultural Research, 72(3), 93, 1946, Department of Agriculture.

\bibitem{sanchez1967flame} Sánchez Tarifa, Carlos and Muñoz Torralbo, Antonio, Flame propagation along the interface between a gas and a reacting medium, 1967, Combustion Institute, Elsevier.

\bibitem{rothermel1972mathematical} Rothermel, Richard C, A mathematical model for predicting fire spread in wildland fuels, 115, 1972, Intermountain Forest \& Range Experiment Station, Forest Service, US.

\bibitem{anderson1981aids} Anderson, Hal E, Aids to determining fuel models for estimating fire behavior, 122, 1981, US Department of Agriculture, Forest Service, Intermountain Forest and Range.

\bibitem{chuvieco1989application} Chuvieco, Emilio and Congalton, Russell G, Application of remote sensing and geographic information systems to forest fire hazard mapping, Remote Sensing of Environment, 29(2), 147--159, 1989, Elsevier.

\bibitem{blackard1999comparative} Blackard, Jock A and Dean, Denis J, Comparative accuracies of artificial neural networks and discriminant analysis in predicting forest cover types from cartographic variables, Computers and Electronics in Agriculture, 24(3), 131--151, 1999, Elsevier.

\bibitem{baba2024neural} Baba, Abdullatif, Neural networks from biological to artificial and vice versa, Biosystems, 235, 105110, 2024, Elsevier.

\bibitem{stojanova2006learning} Stojanova, Daniela and Panov, Panče and Kobler, Andrej and Džeroski, Sašo and Taškova, Katerina, Learning to predict forest fires with different data mining techniques, Conference on Data Mining and Data Warehouses (SiKDD 2006), Ljubljana, Slovenia, 255--258, 2006, Citeseer.

\bibitem{as2023analysis} AS, Remya Ajai and Harikrishnan, Nellippallil Balakrishnan and Nagaraj, Nithin, Analysis of logistic map based neurons in neurochaos learning architectures for data classification, Chaos, Solitons \& Fractals, 170, 113347, 2023, Elsevier.

\bibitem{san2020advance} San-Miguel-Ayanz, Jesús and Durrant, Tracy and Boca, Roberto and Liberta, Giorgio and Branco, Alfredo and De Rigo, Davide and Ferrari, Pieralberto Maianti and Artes Vivancos, Tomas and Costa, Hugo and others, Advance EFFIS report on Forest Fires in Europe, Middle East and North Africa 2017, 2020.

\bibitem{van1974structure} Van Wagner, Charles E and others, Structure of the Canadian forest fire weather index, 1333, 1974, Environment Canada, Forestry Service, Ottawa, ON, Canada.

\bibitem{boser1992training} Boser, Bernhard E and Guyon, Isabelle M and Vapnik, Vladimir N, A training algorithm for optimal margin classifiers, Proceedings of the Fifth Annual Workshop on Computational Learning Theory, 144--152, 1992.

\bibitem{breiman2001random} Breiman, Leo, Random forests, Machine Learning, 45(1), 5--32, 2001, Springer.

\bibitem{quinlan1986induction} Quinlan, J. Ross, Induction of decision trees, Machine Learning, 1(1), 81--106, 1986, Springer.

\bibitem{schapire2013explaining} Schapire, Robert E, Explaining AdaBoost, Empirical Inference: Festschrift in Honor of Vladimir N. Vapnik, 37--52, 2013, Springer.

\bibitem{breiman1996bagging} Breiman, Leo, Bagging predictors, Machine Learning, 24(2), 123--140, 1996, Springer.

\bibitem{chen2016xgboost} Chen, Tianqi and Guestrin, Carlos, XGBoost: A scalable tree boosting system, Proceedings of the 22nd ACM SIGKDD International Conference on Knowledge Discovery and Data Mining, 785--794, 2016.

\end{thebibliography}
\end{document}